\documentclass[sigconf]{acmart}
\AtBeginDocument{%
  \providecommand\BibTeX{{%
    \normalfont B\kern-0.5em{\scshape i\kern-0.25em b}\kern-0.8em\TeX}}}

\copyrightyear{2022}
\acmYear{2022}
\setcopyright{acmcopyright}\acmConference[MM '22]{Proceedings of the 30th ACM International Conference on Multimedia}{October 10--14, 2022}{Lisboa, Portugal} \acmBooktitle{Proceedings of the 30th ACM International Conference on Multimedia (MM '22), October 10--14, 2022, Lisboa, Portugal}
\acmPrice{15.00}
\acmDOI{10.1145/3503161.3547779}
\acmISBN{978-1-4503-9203-7/22/10}

\usepackage{multirow}
\usepackage{diagbox}
\usepackage{makecell}
\definecolor{grey}{RGB}{128,138,135}
\usepackage{bbding}
\usepackage{colortbl}
\usepackage{algorithm}
\usepackage{algorithmic}

\begin{document}

\title{Grouped Adaptive Loss Weighting for Person Search}


\author{Yanling Tian}
\authornotemark[2]
\affiliation{
  \institution{Nanjing University of Science and Technology}
  \country{China}
}
\email{yl.tian@njust.edu.cn}

\author{Di Chen}
\authornotemark[2]
\affiliation{
  \institution{Nanjing University of Science and Technology}
  \country{China}
}
\email{dichen@njust.edu.cn}

\author{Yunan Liu}
\affiliation{
  \institution{School of Artificial Intelligence, Dalian Maritime University}
  \country{China}
}
\email{liuyunan@njust.edu.cn}

\author{Shanshan Zhang}
\authornote{The corresponding author is Shanshan Zhang.}
\authornotemark[2]
\affiliation{
  \institution{Nanjing University of Science and Technology}
  \country{China}
}

\email{shanshan.zhang@njust.edu.cn}

\author{Jian Yang}
\authornote{Yanling Tian, Di Chen, Shanshan Zhang and Jian Yang are with PCA Lab, Key Lab of Intelligent Perception and Systems for High-Dimensional Information of Ministry of Education, and Jiangsu Key Lab of Image and Video Understanding for Social Security, School of Computer Science and Engineering, Nanjing University of Science and Technology.}
\affiliation{%
  \institution{Nanjing University of Science and Technology}
  \country{China}
}
\email{csjyang@njust.edu.cn}

\renewcommand{\shortauthors}{Yanling Tian et al.}

\begin{abstract}
Person search is an integrated task of multiple sub-tasks such as foreground/background classification, bounding box regression and person re-identification. Therefore, person search is a typical multi-task learning problem, especially when solved in an end-to-end manner. Recently, some works enhance person search features by exploiting various auxiliary information, {\it e.g.} person joint keypoints, body part position, attributes, {\it etc.}, which brings in more tasks and further complexifies a person search model. The inconsistent convergence rate of each task could potentially harm the model optimization. A straightforward solution is to \emph{manually assign different weights} to different tasks, compensating for the diverse convergence rates. However, given the special case of person search, {\it i.e.} with a large number of tasks, it is impractical to weight the tasks manually. To this end, we propose a Grouped Adaptive Loss Weighting (GALW) method which adjusts the weight of each task automatically and dynamically. Specifically, we group tasks according to their convergence rates. Tasks within the same group share the same learnable weight, which is dynamically assigned by considering the loss uncertainty. Experimental results on two typical benchmarks, CUHK-SYSU and PRW, demonstrate the effectiveness of our method.
\end{abstract}

\maketitle

\section{Introduction}

\begin{figure}[!t]
\setlength{\abovecaptionskip}{1mm} 
  \centering
  \includegraphics[width=0.9\linewidth]{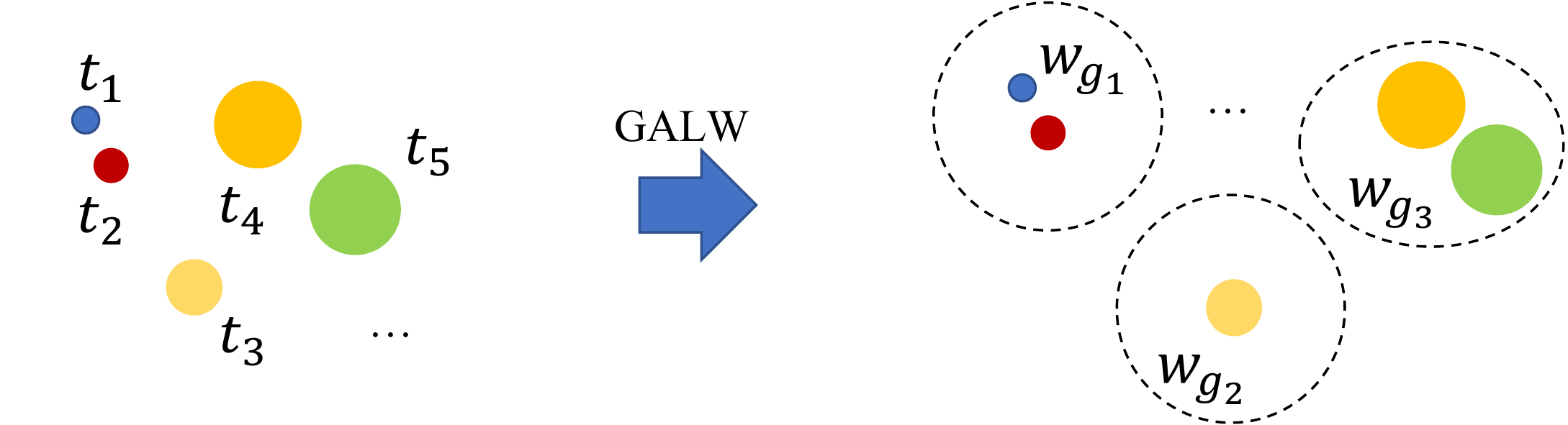}
  \caption{Illustration of our proposed GALW. Different sub-tasks are presented as bubbles in different colors whose area is proportional to the convergence rate of a sub-task in an end-to-end model. We partition a collection of tasks with a consistent convergence rate into a group shared with the same learnable loss weight which is dynamically assigned by considering the loss uncertainty.}
  \Description{...}
  \label{fig:teaser}
\end{figure}

Person search has attracted the attention of many researchers in recent years, as it can be widely applied in video surveillance, social entertainment, {\it etc.} Person search is an integrated task of pedestrian detection and person re-identification (re-id). The goal of this task is to find the given person from a set of uncropped scene images.

Recent works on person search can be roughly divided into one-stage methods~\cite{CUHK,NPSM,RCAA,CTXG,QEEPS,HOIM,BINet,NAE,PGA,SeqNet,TriNet,Alignps,Alignps_roi,yang2020bottom,zhang2021boosting,liu2020dual} and two-stage methods~\cite{PRW,CNN+MGTS,CNN+CLSA,FPN+RDLR,IGPN,OR,TCTS}. A typical two-stage method first trains a pedestrian detection model and then applies a re-id model on the detected pedestrian images. In contrast, one-stage methods optimize pedestrian detection and re-id tasks simultaneously in an end-to-end manner. To increase the discrimination ability of models, some works introduce auxiliary tasks~\cite{Keypoint,CNN+MGTS,TriNet,part} ({\it e.g.} pose estimation, attribute recognition, and human parsing) to provide guidance information for person search. Han \emph{et al.}~\cite{TriNet} adopt part classification to obtain spacial fine-grained features. Zhong \emph{et al.}~\cite{part} extract part features from visible body parts and take re-id as a partial feature matching procedure. Chen \emph{et al.}~\cite{CNN+MGTS} explore the impact of background information on person search by semantic segmentation. Although auxiliary information brings an improvement on performance, jointly optimizing such model containing multiple tasks becomes complicated, which is mainly caused by the inconsistent convergence rates of different tasks. Therefore, it is a great challenge to design an effective multi-task learning (MTL) strategy for model optimization. 

To synchronize the convergence of different tasks, a straightforward solution is to manually assign different loss weights for these tasks. However, it is difficult to assign a suitable weight for each task manually. The automatic loss weighting strategy~\cite{Uncertainty,DWA,GradNorm,DTP} provides an alternative way to solve this problem. Kendall \emph{et al.}~\cite{Uncertainty} propose a novel multi-task loss that uses homoscedastic uncertainty to weight tasks dynamically. Liu \emph{et al.}~\cite{DWA} use dynamic weight averaging to balance learning speed of each task. All these methods have achieved good results in MTL. However, through analysis of the impact of loss weighting on person search, we find that the performance degrades when optimizing too many tasks in an end-to-end manner.

To solve this problem, in this work, we adopt the task grouping strategy, which assembles many tasks into a small number of optimization groups. 
Existing methods~\cite{MTL_S2021,MTL_S2020,Group20,Group21} have shown the efficacy of grouping different tasks, {\it e.g.} Fifty \emph{et al.}~\cite{Group21} determine task groups by employing a measure of inter-task affinity. However, these methods are inconvenient, time and resource consuming since task groups are usually associated with \textbf{different} networks which require separate training.
Different from these methods, as shown in Fig.~\ref{fig:teaser}, we propose a grouped adaptive loss weighting (GALW) method that groups tasks according to their convergence rates in the \textbf{same} network. Specifically, we put those tasks with a similar gradient magnitude slope into a group and explore homoscedastic uncertainty learning to optimize the weights of different grouped tasks for person search. By doing this, we can dynamically learn the optimal loss weights, which makes optimization more effective and stable without extra computational costs. To verify the extendability of our method on more tasks, we additionally employ an attribute recognition network that obtains rich features benefit for the problem of mismatch on similar appearances. 

\begin{figure*}
\setlength{\abovecaptionskip}{0mm} 
  \includegraphics[width=0.88\textwidth]{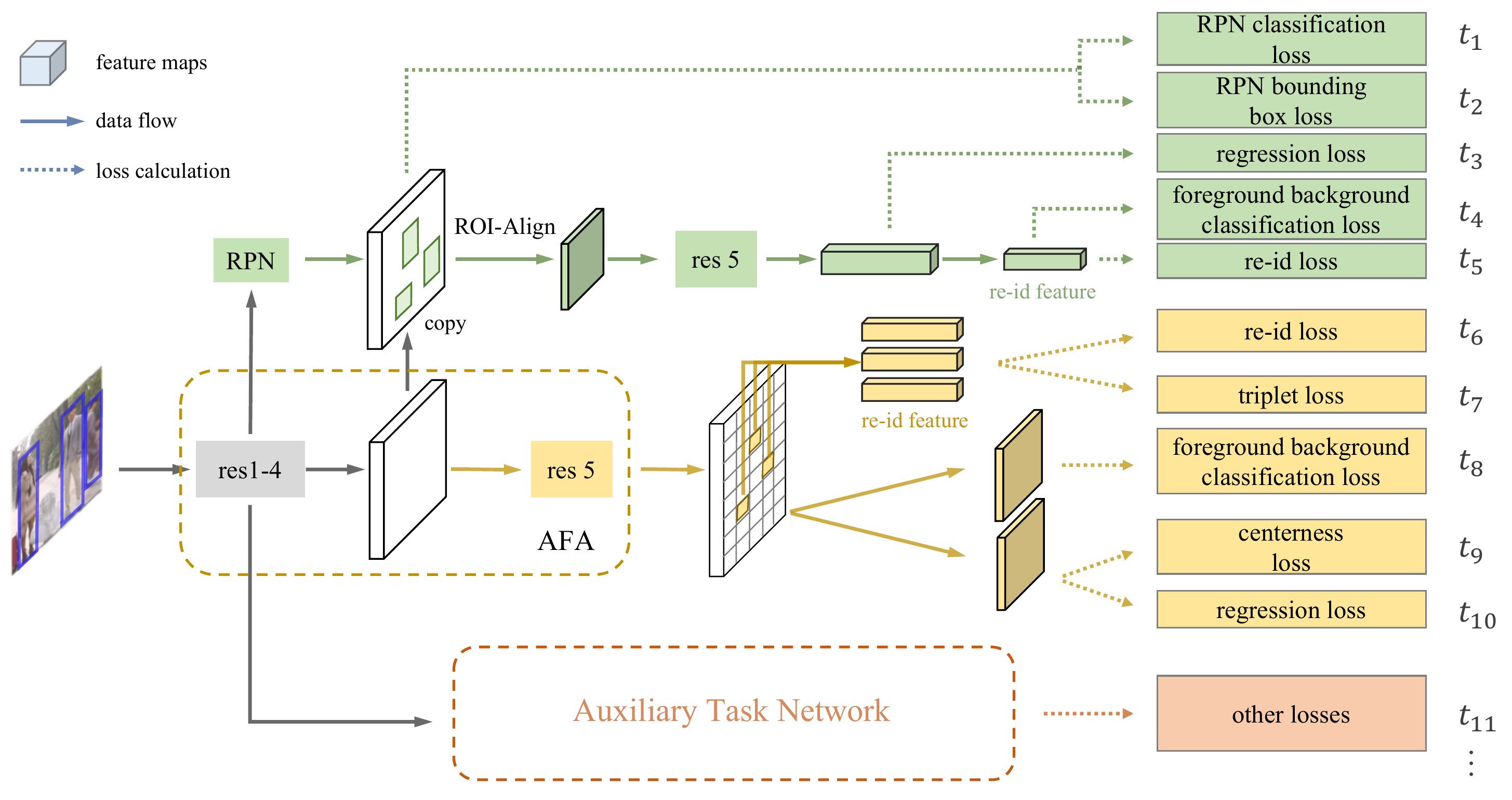}
  \caption{The overview of our network and loss functions. This framework uses ResNet-50~\cite{ResNet} as backbone and contains three branches, NAE branch~\cite{NAE} (green part), AlignPS branch~\cite{Alignps} (yellow part), and auxiliary task branch (orange part), of which the first two branches come from ROI-AlignPS~\cite{Alignps_roi}. $t_k$ in the right is the corresponding loss function in the box of task $k$. We can extend model by adding auxiliary tasks. AFA means aligned feature aggregation module used in~\cite{Alignps} and RPN~\cite{fasterrcnn} refers to region proposal network.}
  \Description{...}
  \label{fig:network}
\end{figure*}

In summary, our contributions are as follows:

  \begin{enumerate}
  
    \item We provide an analysis to explore the effect of uncertainty loss weighting strategy on person search. We find that the performance degrades when dealing with large numbers of different tasks in person search and the issue of inconsistent convergence rates gets severe as the number of tasks grows.
    \item  We propose a grouped adaptive loss weighting (GALW) method, which adjusts the weight of each task automatically and dynamically. We put those tasks with similar convergence rates into a group shared with the same learnable loss weight, which makes the model optimization more effective for person search.
    \item We achieve state-of-the-art results on the CUHK-SYSU dataset and competitive performance with less running time on the PRW dataset. Furthermore, we verify the generalization ability and extendability of GALW by applying GALW on different baselines and adding auxiliary tasks respectively.

\end{enumerate}


\section{Related Work}
In this section, we first review existing works on person search. Since person search is a typical MTL problem, we also review some related works about MTL. 
\subsection{Person Search}

The task of person search traditionally consists of two sub-tasks: pedestrian detection and re-id. Existing methods can be categorized into one-stage or two-stage models according to their training strategy (separately or end-to-end). DPM+IDE~\cite{PRW} is the first two-stage framework that combines different detectors and re-id models to detect pedestrians first and then perform re-id using the cropped images. Based on that, several methods make further improvements and achieve better performance~\cite{TCTS,OR,IGPN,FPN+RDLR,CNN+MGTS,CNN+CLSA}. These two-stage models lack efficiency although they have better performance.

For one-stage methods, Xiao \emph{et al.}~\cite{CUHK} first propose an end-to-end framework for person search by adding re-id layers after Faster-RCNN~\cite{fasterrcnn}. Chen \emph{et al.}~\cite{NAE} introduce a norm-aware embedding method (NAE) which relieves the contradictory goals of pedestrian detection and person re-id tasks by decomposing the feature embedding into norm and angle respectively. Based on that, SeqNet~\cite{SeqNet} gets better performance by stacking the NAE models.  Kim \emph{et al.}~\cite{PGA} present a prototype-guided attention module to obtain discriminative re-id features. Yan \emph{et al.}~\cite{Alignps} first propose the anchor-free based person search model (AlignPS) which addresses the problems of scale, region and task misalignment. They further introduce an advanced AlignPS (ROI-AlignPS~\cite{Alignps_roi}), taking advantage of the anchor-based and anchor-free model, to enhance the final performance.

Recently, some one-stage works enhance person search features by exploiting various auxiliary information. Chen \emph{et al.}~\cite{Keypoint} explore skeleton key points to update spatial-temporal features. Han \emph{et al.}~\cite{TriNet} incorporate the part classification branch to generalize features shared with pedestrian detection and re-id and further enhance the quality of spacial features according to the detection confidence, as well as preventing the detection over-fitting in the latter part of the training.

Although these one-stage methods are simple and efficient, they assign different weights for different tasks manually and do not take model optimization into account. Especially with more tasks, there is a non-negligible problem that how to balance the contributions of losses for optimizing an end-to-end model. For two-stage models, it is unnecessary to focus on this problem and there is no need to optimize all losses together because they train the detector and re-id separately. Since person search is a typical MTL problem and loss weighting is a straightforward way to optimize models by re-weighing losses during training, in this work, we make an analysis on loss weighing with different numbers of tasks in the field of person search, which further helps us optimize this task. 

\subsection{Multi-Task Learning}
MTL is a machine learning method that is widely used in many fields such as computer vision~\cite{DWA,sharedtrunk,sharedtrunk14}, reinforcement learning~\cite{RL1}, natural language processing~\cite{NLP1,NLP2}. In MTL, multiple tasks can be trained simultaneously and the loss functions can be optimized at once within a single model. According to the existing frameworks, MTL methods can be divided into three aspects: architecture design~\cite{sharedtrunk,sharedtrunk14}, optimization strategy ~\cite{Uncertainty,DWA,GradNorm,DTP} and task relationship learning~\cite{MTL_S2021,MTL_S2020}. 

\textbf{Architecture designing based methods} focus on which components can be shared and which are task-specific to get generalized features for each task~\cite{DWA,sharedtrunk,sharedtrunk14}. \textbf{Optimization strategy based methods} aim to solve the task balancing problem during training to optimize the model parameters in a faster learning speed. One of the most common methods is loss weighting~\cite{Uncertainty,DWA,GradNorm,DTP}. Yan \emph{et al.}~\cite{Uncertainty} use homoscedastic uncertainty in Bayesian modeling to weight losses which makes it more suitable for noisy data. Chen \emph{et al.}~\cite{GradNorm} propose gradient normalization to balance learning speed and magnitude of different losses but need to calculate the gradient which needs more GPU resources and increases training time. Guo \emph{et al.}~\cite{DTP} present a dynamic task prioritization method that prefers hard-learning tasks. \textbf{Task relationship learning based methods} aim to learn the relationship of tasks and further improve the learning on these tasks with learned relationships. Task grouping is the most typical method through different ways. Standley \emph{et al.}~\cite{Group20} make an analysis on the influence factors of MTL and get better performance under a limited inference-time budget by using a network selection strategy. Fifty \emph{et al.}~\cite{Group21} propose a new measure of inter-task affinity to group tasks by quantifying the effect between tasks and train the grouped tasks separately. 

These loss weighting methods optimize training problems in terms of learning speed, performance, uncertainty and order of loss magnitude. Meanwhile, task grouping provides us with a way to learn an explicit representation of tasks or relationships between tasks. Different from the above techniques, we propose a grouped adaptive loss weighting method that exploits the advantages of both. We regard the similar gradient magnitude trend as consistent convergence rate and partition a collection of tasks with similar gradient magnitude trend into a group. 
We further exploit homoscedastic uncertainty learning to assign loss weights on the different groups and improve model performance without extra computational costs.

\section{Method}
\begin{figure*}[!t]
\setlength{\abovecaptionskip}{0.88mm} 
\centering
\includegraphics[width=0.9\linewidth]{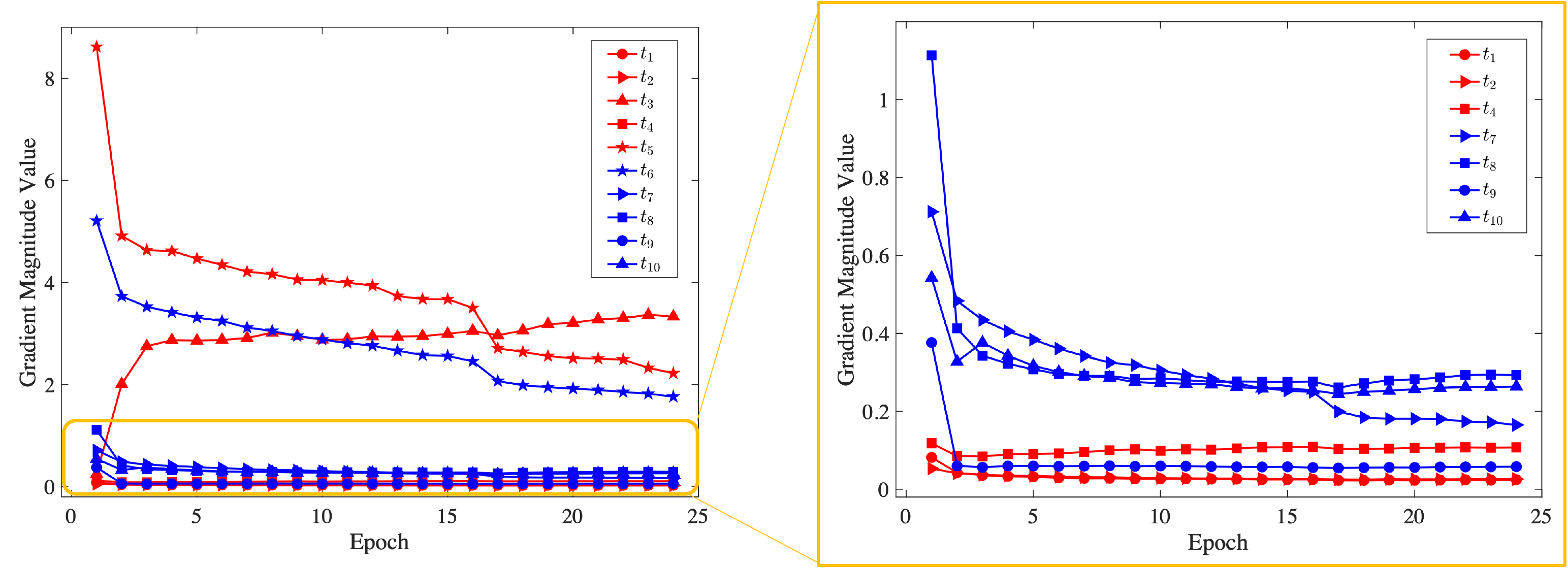}
\caption{Gradient magnitudes of different sub-tasks in ROI-AlignPS$^\ast$. We zoom the lower part of the left figure (orange solid rectangle) for better view.}
\label{fig:grad_alignps_prw}
\end{figure*}

In this section, we first make an analysis on loss weighting to demonstrate our motivation, followed by the proposed method for person search. 

\subsection{Analysis on Loss Weighting}
\label{sec:analysis}

Person search is a typical MTL problem, however, some of the existing MTL strategies are not suitable for optimizing a person search model which contains too many sub-tasks. In this subsection, we conduct a comprehensive analysis of loss weighting strategy on person search. Specifically, We use two recent works, i.e. AlignPS~\cite{Alignps} and ROI-AlignPS~\cite{Alignps_roi}, as our baseline methods. As shown in Fig.~\ref{fig:network}, ROI-AlignPS contains 10 sub-tasks (i.e. $t_1-t_{10}$), while AlignPS is an anchor-free method which only contains 5 sub-tasks (i.e. $t_6-t_{10}$). Due to the fact that the mutual learning of ROI-AlignPS is not performed in every iteration of the training phase, we redesign ROI-AlignPS (denoted as ROI-AlignPS$^\ast$) by removing mutual learning loss. We use uncertainty loss~\cite{Uncertainty} as our loss weighting function, (denotes uncertainty loss weighting function (ULWF) in the following), which uses homoscedastic uncertainty to automatically adjust weights. Compared to other loss weighting methods, uncertainty loss is easy to implement without any gradient calculations and does not introduce too many extra parameters.

We conduct experiments on one of typical person search datasets, PRW~\cite{PRW} dataset. The mean average precision (mAP) and Top-1 are used as the evaluation metrics, a higher value indicates a better performance of person search. From Fig.~\ref{fig:grad_alignps_prw}, Tab.~\ref{table:weighting} and Tab.~\ref{table:relation}, we make the following observations and discussions:

\setlength{\tabcolsep}{2.5pt}
\begin{table}[!t]
\begin{center}
\caption{Effect of loss weighting on performance of ROI-AlignPS$^\ast$~\cite{Alignps_roi} and AlignPS~\cite{Alignps} when their loss of sub-task $t_i$ is manually assigned different weight $w_i$ ($i=1,2,...,10$).}
\label{table:weighting}
\begin{tabular}{ccccccccccccc}
\hline\noalign{\smallskip}
Method& $w_1$ & $w_2$ & $w_3$ & $w_4$ & $w_5$ & $w_6$ & $w_7$ & $w_8$ & $w_9$ & $w_{10}$ & mAP & Top-1\\
\noalign{\smallskip}
\hline
\noalign{\smallskip}
\multirow{4}{*}{\rotatebox{90}{AlignPS}}&
       - & - & - & - & - & 1 & 1 & 1 & 1 & 1     & 45.90  & 81.90\\
  &     - & - & - & - & - & 10& 1 & 1 & 1 & 10    & \textbf{22.51}  & \textbf{62.22}\\
   &    - & - & - & - & - & 1 & 1 & 10& 1 & 10     & 39.03  & 79.21\\
   &    - & - & - & - & - & 1 & 1 & 10& 10& 1     & \textbf{47.83}  & \textbf{81.88}\\
   
 &&&&&&&&&&&&  \\
 \multirow{5}{*}{\rotatebox{90}{ROI-AlignPS$^\ast$ }}&    1 & 1 & 1 & 1 & 1 & 1 & 1 & 1 & 1 & 1     & 48.68  & 84.15\\
 &    10& 10& 1 & 10& 1 & 1 & 10& 1 & 1 & 10     & 47.45  & 82.47\\
 &    10& 1 & 10& 1 & 10& 1 & 10& 10& 10& 1     & \textbf{46.41}  & \textbf{81.88}\\
 &    1 & 1 & 1 & 1 & 10& 10& 10& 10& 10& 1     & 49.79  & 83.65\\
 &    1 & 1 & 10 & 1 & 10& 10& 10& 10& 10& 1     & \textbf{50.30}  & \textbf{84.30}\\

\hline
\end{tabular}
\end{center}
\end{table}
\setlength{\tabcolsep}{1.4pt}

\setlength{\tabcolsep}{1.4pt}
\begin{table}[!t]
\begin{center}
\caption{Effectiveness of ULWF on different number of tasks.}
\label{table:relation}
\begin{tabular}{lccc}
\hline\noalign{\smallskip}
Method & Task number & mAP & Top-1\\
\noalign{\smallskip}
\hline
\noalign{\smallskip}
AlignPS~\cite{Alignps}                     & 5  & 45.58  & 81.90\\
AlignPS w/ ULWF              & 5  & 48.88 & 82.91\\
\\
ROI-AlignPS$^\ast$~\cite{Alignps_roi}          & 10  & 50.30  & 84.30\\
ROI-AlignPS$^\ast$ w/ ULWF    & 10  & 45.76 & 82.77\\

\hline
\end{tabular}
\end{center}
\end{table}
\setlength{\tabcolsep}{1.4pt} 

\begin{enumerate}

    \item Fig.~\ref{fig:grad_alignps_prw} depicts gradient magnitude of different tasks in ROI-AlignPS$^\ast$ during training. We can see that there are diverse convergence rates in different tasks. Convergence rates of $t_3$ and others show opposite trends over training.
  
    \item Tab.~\ref{table:weighting} shows the performance of AlignPS and ROI-AlignPS$^\ast$ with manually routine assigned with loss weights. We find that the performance gaps of ROI-AlignPS and AlignPS are large when their sub-tasks are manually assigned different weights. For example, the best performance of AlignPS outperforms its worst performance by 25.32 pp w.r.t mAP and 19.66 pp w.r.t Top-1. For the MTL in person search, assigning a suitable weight for each task is important but is hard to be achieved manually.

    \item Tab.~\ref{table:relation} explores the effectiveness of ULWF. Compared to the baseline methods AlignPS, using ULWF brings a significant improvement of 3.30\% w.r.t mAP. When using ULWF, the mAP of ROI-AlignPS decreases by 4.54\%. We find that ULWF is useful for learning a small number of tasks, however, it causes performance degradation when optimizing too many tasks in an end-to-end manner. We consider that the inconsistent convergence rates of too many tasks lead to the under-performance of the loss weighting strategy.
    
\end{enumerate}

\subsection{Grouped Adaptive Loss Weighting}
\label{sec:method}

Based on the analysis in Sec.~\ref{sec:analysis}, we propose a grouped adaptive loss weighting method. In the following, we first introduce our regularized uncertainty loss weighting function (RULWF) in Sec.~\ref{sec:method1.1}. Then the details of task grouping are described in Sec.~\ref{sec:method1.2}.

\subsubsection{Regularized Uncertainty Loss Weighting}
\label{sec:method1.1}
 We use homoscedastic uncertainty learning~\cite{gal2016dropout,kendall2017uncertainties,Uncertainty} to conduct loss weighting. Take classification task as an example. The Bayesian probabilistic likelihood of model output is defined as:
\begin{align}
  p(y|f(x,W),\sigma_i) = Softmax(\frac{1}{\sigma_{i}^{2}}f(x,W)),\label{equ:pro}
\end{align}
where $f(x,W)$ is the output of the network and $\sigma_i$ is the observation noise of task $i$ which is a learnable parameter. The log-likelihood of this output is:
\begin{equation}
\begin{aligned}
      \log(p(y=c|f(x,W),\sigma_i)) &= \frac{1}{\sigma_{i}^{2}}f_{c}(x,W) \\
  &- \log\sum_{j=0}^{C}exp(\frac{1}{\sigma_{i}^{2}}f_{j}(x,W)),\label{equ:likelihood}
\end{aligned}
\end{equation}
where $C$ is the number of classes. And the loss function can be written as:
\begin{equation}
\begin{aligned}
  L_i(x,W,\sigma_i)&=-\log(p(y|f(x,W),\sigma_i)) \\
                 &=\frac{1}{\sigma_{i}^{2}}L_{ori_{i}}(x,W)+\frac{\sum_{j=0}^{C}exp(\frac{1}{\sigma_{i}^{2}}f_{j}(x,W))}{(\sum_{j=0}^{C}exp(f_{j}(x,W) ))^{\frac{1}{\sigma_{i}^{2}}}}\\
                 &\approx \frac{1}{\sigma_{i}^{2}}L_{ori_{i}}(x,W)+\log\sigma_i,\label{equ:l}
\end{aligned}
\end{equation}
where $L_{ori_{i}}(x,W)=-\log(p(y=c|f(x,W)))$ denotes original unweighted loss function of task $i$. In order to simplify the optimisation objective, an assumption $\frac{1}{\sigma_{i}^{2}}\sum_{j=0}^{C}exp(\frac{1}{\sigma_{i}^{2}}f_{j}(x,W))=(\sum_{j=0}^{C}exp(f_{j}(x,W) ))^{\frac{1}{\sigma_{i}^{2}}}$ is used in the last transition and it becomes equality when $\sigma_i=1$~\cite{Uncertainty}.

Furthermore, some changes need to be made to meet that assumption and stabilize the training. We add a regularizer loss for each task:
\begin{align}
  L_{regularizer_{i}} = || \sigma_i-1||_1,\label{equ:reg}
\end{align}
where $||.||_1$ is $l$1-norm. Then, the total loss is written as:
\begin{equation}
\begin{aligned}
L_{total}&=L_{tasks}+\lambda L_{regularizer} \\
        &=\sum_{i=0}^{T} L_i(x,W,\sigma_i)+\lambda \sum_{i=0}^{T} L_{regularizer_i} \\
        &=\sum_{i=0}^{T} \left ( \frac{1}{k\sigma_{i}^{2}} L_{ori\_i}(x, W) + \log\sigma_i + \lambda || \sigma_i-1||_1   \right ),\label{equ:gen}
\end{aligned}
\end{equation}
in which $k=1,2$ when the task is classification task and regression task respectively. $\lambda$ is the weight for the regularizer term. In practice, we set $\lambda=1$. We will validate the effectiveness of this design in Sec.~\ref{sec:ablation}.

\subsubsection{Task Grouping}
\label{sec:method1.2}

As discussed in Sec.~\ref{sec:analysis}, when the number of tasks increases, the relationship between tasks becomes particularly intricate. Task grouping is one way of learning relationships between tasks and leverage the learned task relationships to improve model training.
We measure the gradient slope of each tasks and group them according to the similarity of gradient slopes, which we deem is an effective strategy for multi-task learning.

Specifically, we calculate the gradient magnitude of shared parameters at each epoch:
\begin{align}
  \Gamma = \frac{1}{M}\sum_{m=1}^{M} norm(\theta_m),\label{equ:mag}
\end{align}
where $M$ is the number of shared parameters. We use the average slope of each polyline in Fig.~\ref{fig:grad_alignps_prw} as a measure of task trends.
\begin{align}
  s_i = \frac{1}{N-1}\sum_{n=1}^{N-1} (\Gamma_{N+1} - \Gamma_N),\label{equ:slope}
\end{align}
where $N$ is the number of epochs. $s_i$ refers to the average slope of task $i$. In general, the convergence rate decreases and gradually plateaus over time. The average slope for each task is too small to distinguish, so we process it further as follows:
\begin{align}
  s_i^\ast=sign(s_i) \xi (\log |s_i|),\label{equ:log}
\end{align}
where $sign(.)$ and $\log(.)$ are used to indicate the sign and the order of magnitude of $s_i$ respectively. $\xi(.)$ is \textit{sigmoid} activation function which makes it insensitive to numbers that orders of magnitude are large. Then, we partition $s^\ast=\{s_1^\ast,s_2^\ast,...,s_T^\ast\}$ into different groups by hierarchical clustering algorithm~\cite{HA}. And final loss function Eq.~(\ref{equ:gen}) turns into:
\begin{align}
  L_{total} = \sum_{g=0}^{G} L_g(x,W,\sigma_g)+\lambda L_{regularizer} ,\label{equ:final}
\end{align}
where $G$ refers to the number of groups and loss function of $g^{th}$ group equals to the sum of the losses of tasks within this group. We describe the detailed procedure in Algorithm~\ref{alg:algorithm1}.

\begin{algorithm}[!t]
\begin{algorithmic}[1]
\caption{Grouped Adaptive Loss Weighting Method} %
\label{alg:algorithm1}
\REQUIRE image data $X$ from dataset. \textit{Initialise:} Model shared parameters $\Theta^\ast$ in backbone, epoch number $N$, task number $T$, group number $G$;  \\ 
\ENSURE Model parameters $\Theta$; \\ 

    \STATE \textcolor{blue}{// For first training;}
    \FOR{$n=1,2,...,N$} 
        \FOR{$t=1,2,...,T$}
        
            \STATE Calculate gradient magnitude $\Gamma_{n}^{t}$ of $\Theta^\ast$ using Eq.~(\ref{equ:mag});
        \ENDFOR
    \ENDFOR
    
    \FOR{$t=1,2,...,T$}
        \STATE Gradient magnitude trend of task: $\{\Gamma_{1},\Gamma_{2},...,\Gamma_{N}\}$
        \STATE Calculate average slope $s_t$ using Eq.~(\ref{equ:slope}) and get $s^\ast$ by processing $s_t$ with Eq.~(\ref{equ:log})
    \ENDFOR
    \STATE Task grouping with hierarchical clustering using \\ $s^\ast=\{s_1^\ast,s_2^\ast,...,s_T^\ast\}$.
    \STATE \textcolor{blue}{// For second training;}
    \STATE Grouping tasks: $Group_1, Group_2,..., Group_g$;
    \STATE Training Model with grouped tasks using Eq.~(\ref{equ:final});\\
    \RETURN Final model parameters $\Theta$;
\end{algorithmic}
\end{algorithm}

\section{Experiments}

In this section, we first introduce two typical datasets, CUHK-SYSU~\cite{CUHK} and PRW~\cite{PRW}, evaluation metrics of person search, followed by implementation details. Then we perform exhaustive ablation studies on PRW to examine the contributions of the proposed components. Finally, we show the experimental results in comparison to state-of-the-art methods.

\subsection{Datasets and Settings}
There are two typical datasets in person search, which we use in our experiments. 

(1) \textbf{CUHK-SYSU.}  
CUHK-SYSU~\cite{CUHK} is a large-scale person search database providing 18,184 images, 96,143 pedestrian bounding boxes and 8,432 different identities. There are two kinds of images: video frames selected from movie snapshots and street/city scenes captured by a moving camera. The dataset is divided into standard train/test split, where the training set includes 11,206 images and 5,532 identities, while the testing set contains 6,978 gallery images with 2,900 query persons. Instead of using all the test images as a gallery, it defines a set of protocols with gallery sizes ranging from 50 to 4,000. We use the default gallery size of 100 in our experiments unless otherwise specified.

(2) \textbf{PRW}. 
PRW dataset~\cite{PRW} contains 11,816 frames which are sampled from videos captured with six cameras on a university campus. There are 43,110 bounding boxes and where 34,304 of them are annotated with 932 identities and the rest marked as unknown identities. In the training set, it provides 5,134 images with 482 different persons while the testing set includes 2,057 query persons and 6,112 gallery images. Different from CUHK-SYSU, we use the whole gallery set as the search space for each query person. 

\textbf{Evaluation Metrics}. 
Similar to re-id, we employ mAP and cumulative matching characteristics  Top-$k$ as performance metrics for person search, where mAP metric reflects the accuracy and matching rate of searching a probe person from gallery images. Top-$k$ score represents the percentage that at least one of the $k$ proposals most similar to a given query succeeds in the re-id matching.

\subsection{Implementation Details}
Our model consists of three branches as shown in Fig.~\ref{fig:network}: NAE branch, AlignPS branch and auxiliary task branch. The first two of them make up our baseline ROI-AlignPS$^\ast$~\cite{Alignps_roi}. In addition, we add an attribute recognition network branch as an auxiliary task as an example to verify the extendability of our proposed method on more tasks, which is denoted as ROI-AlignPS$^\ast$-Attr. We extract different level features of proposals which are cropped and reshaped to $32\times32$ in pyramid structure and put them into attribute localization module (ALM) modules~\cite{Attribute} to perform attribute prediction. Binary cross-entropy loss function is adopted as attribute loss $t_{11}$. Due to lack of attribute labels on CUHK-SYSU or PRW dataset, we use an off-the-shelf framework~\cite{Attribute} to generate attribute pseudo-labels and directly conduct this supervised learning. All frameworks employ ResNet-50~\cite{ResNet} pretrained on ImageNet~\cite{Imagenet} as the backbone, where blocks from ``conv2'' to ``conv4'' are the shared layers.

During training, all experiments are conducted on a single NVIDIA TITAN RTX GPU for 24 epochs with an initial learning rate 0.001, which is reduced by 10 at epoch 16 and 22 respectively. Following~\cite{Alignps_roi}, the momentum and weight decay of stochastic gradient descent (SGD) optimizer are set to 0.9 and 0.0005. Warmup strategy is used for 300 steps. 
We adopt a multi-scale training strategy where the long sides of images are randomly resized from 667 to 2000 during training. At test time, the scale of the test image is fixed into $1500\times900$. Our code will be made publicly available and more details are provided in our code.

\subsection{Ablation Study}
\label{sec:ablation}
In this section, we conduct several analytical experiments on PRW dataset~\cite{PRW} to better understand our proposed method.

\subsubsection{\textbf{Necessity of loss weighting and task grouping.}} In order to verify the effect of loss weighting and task grouping in person search, we perform several experiments. As introduced in Sec.~\ref{sec:method}, our GALW consists of a loss weighting function RULWF without task grouping and a task grouping strategy. We manually delete one of them and show their performance in Tab.~\ref{table:grouping}. This table shows that (1) our RULWF outperforms ULWF both with and without task grouping (row 2 \& 3 and row 4 \& 5) (2) When we add our task grouping method either on ULWF or RULWF, performance has been significantly improved, which means our task grouping method is beneficial for training. (3) The gap between the first and last rows demonstrates that our proposed method GALW is an effective way to dynamically weight tasks.

\setlength{\tabcolsep}{4pt}
\begin{table}[!t]
\begin{center}
\caption{The performance of our method GALW with different components. ULWF means multi-task loss in~\cite{Uncertainty} while RULWF refers to regularized ULWF in our method.}
\label{table:grouping}
\begin{tabular}{ccc|cc}
\hline\noalign{\smallskip}
 ULWF&RULWF &Task Grouping& mAP  & Top-1\\
\noalign{\smallskip}
\hline
\noalign{\smallskip}
&&     &        50.30  & 84.30\\
\Checkmark&&           & 45.76  & 82.77\\
&\Checkmark&            & 50.92  & 85.63\\
\Checkmark&& \Checkmark    & 52.42  & 85.43\\
&\Checkmark& \Checkmark    & \textbf{52.89}  & \textbf{86.07}\\

\noalign{\smallskip}
\hline
\end{tabular}
\end{center}
\end{table}
\setlength{\tabcolsep}{1.4pt}

\setlength{\tabcolsep}{4pt}
\begin{table}[!t]
\begin{center}
\caption{Comparative results on different numbers of groups. GN refers to the number of groups.}
\label{table:grouping_thresh}
\begin{tabular}{cc|cc}
\hline\noalign{\smallskip}
 GN & Group Details& mAP & Top-1\\
\noalign{\smallskip}
\hline
\noalign{\smallskip}
 2    &\{$t_3$\}, \{$t_1, t_2,t_4,t_5,t_6,t_7,t_8,t_9,t_{10}$\}   &51.99   &85.53 \\
 3    &\{$t_3$\}, \{$t_5, t_6$\}, \{$t_1, t_2,t_4,t_7,t_8,t_9,t_{10}$\}   &52.52  &85.78 \\
 4    &\{$t_3$\},\{$t_1, t_2,t_4, t_9$\},\{$t_8, t_{10}$\}, \{$t_5, t_6,t_7$\}       & \textbf{52.89}  & \textbf{86.07}\\

 5    &\{$t_3$\},\{$t_4$\},\{$t_7$\}, \{$t_5, t_6$\}, \{$t_1, t_2,t_8, t_9, t_{10}$\}   & 52.55  &85.43 \\
 6    &\{$t_1, t_2$\},\{$t_3,t_4$\}, \{$t_5, t_6$\}, \{$t_7$\},\{$t_8, t_{10}$\},\{$t_9$\}   & 52.17  &84.48 \\
 -    & -  &50.92  & 85.63 \\

\noalign{\smallskip}
\hline
\end{tabular}
\end{center}
\end{table}
\setlength{\tabcolsep}{1.4pt}

\setlength{\tabcolsep}{4pt}
\begin{table}[!t]
\begin{center}
\caption{Comparative results on different grouping strategies. GS refers to group strategies.}
\label{table:grouping_way}
\begin{tabular}{lc|cc}
\hline\noalign{\smallskip}

 GS & Group Details & mAP & Top-1\\
\noalign{\smallskip}
\hline
\noalign{\smallskip}
 branch& \{$t_1, t_2, t_3, t_4, t_5$\},\{$t_6, t_7, t_8, t_9,t_{10}$\}     & 51.37	  & 85.53\\
\noalign{\smallskip}
\hline
\noalign{\smallskip}
 \multirow{2}{*}{semantic}& \{$t_5, t_6,t_7$\}, \{$t_1, t_2, t_3, t_4, t_8, t_9, t_{10}$\}     & 51.80	 & 85.58\\
    & \{$t_5, t_6,t_7$\}, \{$t_3, t_{10}$\}, \{$t_1, t_2,t_4,t_8,t_9$\}  & 52.16  & 85.73\\

\noalign{\smallskip}
\hline
\noalign{\smallskip}
 \multirow{2}{*}{random}  &\{$t_1,t_5$\},\{$t_3, t_7$\},\{$t_6, t_{10}$\}, \{$t_2, t_4,t_8,t_9$\}       & 52.14  & 85.78\\
   &\{$t_1,t_2$\},\{$t_3,t_5, t_7$\},\{$t_6$\}, \{$ t_4,t_8,t_9, t_{10}$\}       & 49.58  & 84.54\\
\hline
\noalign{\smallskip}

 ours  &\{$t_3$\},\{$t_1, t_2,t_4, t_9$\},\{$t_7,t_8, t_{10}$\}, \{$t_5, t_6$\}       & \textbf{52.89}  & \textbf{86.07}\\
\hline
\noalign{\smallskip}
 -  &-       & 50.92  & 85.63\\

\noalign{\smallskip}
\hline
\end{tabular}
\end{center}
\end{table}
\setlength{\tabcolsep}{1.4pt}

\subsubsection{\textbf{Group number in hierarchical clustering algorithm.}}

We discuss the impact of different numbers of groups in hierarchical clustering algorithm~\cite{HA}. As shown in Tab.~\ref{table:grouping_thresh}, we can see that different numbers of groups have different effects on performance. The best performance is achieved when the tasks are divided into four groups. In addition, no matter what the number of groups is, experiments with our task grouping method perform better than those without.

\subsubsection{\textbf{Grouping strategies in task grouping.}}
In Tab.~\ref{table:grouping_way}, we compare the performance with different grouping strategies: branches, semantic (one is reid and detection, another is reid, bbox regression and others), random and ours. We can find that (1) our grouping method is better than other three grouping strategies. (2) The method of random grouping divides tasks into four groups same as ours but the performance drops 1.34 pp. in mAP compared with the last row without any grouping strategy, which demonstrates the effectiveness of our grouping method.

\setlength{\tabcolsep}{4pt}
\begin{table}[!t]
\begin{center}
\caption{Extendability of our method GALW on more tasks. ROI-AlignPS$^\ast$-Attr denotes our baseline with an attribute recognition task $t_{11}$.}
\label{table:attr}
\begin{tabular}{lcc}
\hline\noalign{\smallskip}
Method  & mAP & Top-1\\
\noalign{\smallskip}
\hline
\noalign{\smallskip}
ROI-AlignPS$^\ast$~\cite{Alignps_roi}       & 50.30  & 84.30\\
ROI-AlignPS$^\ast$-Attr               & 51.42  & 85.23\\
ROI-AlignPS$^\ast$-Attr w/ GALW         & \textbf{53.25}  & \textbf{86.22}\\
\hline
\end{tabular}
\end{center}
\end{table}
\setlength{\tabcolsep}{1.4pt}

\setlength{\tabcolsep}{4pt}
\begin{table}[!t]
\begin{center}
\caption{Generalization ability of our method GALW by applying GALW to different baselines. }
\label{table:gen}
\begin{tabular}{lccc}
\hline\noalign{\smallskip}
Methods &Group Number & mAP & Top-1\\
\noalign{\smallskip}
\hline
\noalign{\smallskip}
NAE~\cite{NAE}  &- & 43.3  & 80.9\\
NAE w/ GALW &3   & 43.6  & 81.0\\

\\
AlignPS~\cite{Alignps}& - & 45.9  & 81.9\\
AlignPS w/ GALW &3 & 49.4  & 83.5\\
\\
ROI-AlignPS$^\ast$~\cite{Alignps_roi} &- & 50.3  & 84.3\\
ROI-AlignPS$^\ast$ w/ GALW & 4 & 52.9  & 86.1\\
 \noalign{\smallskip}
\hline
\end{tabular}
\end{center}
\end{table}
\setlength{\tabcolsep}{1.4pt}

\subsubsection{\textbf{Extendability of our method}}
In order to verify the extendability of our method on more tasks, we directly add an attribute recognition task as an auxiliary task in Fig.~\ref{fig:network}. We make an analysis from two aspects: (1) whether the performance of the network is improved after adding auxiliary tasks, (2) and whether the performance improvement comes from our method. We group these tasks according to their convergence rates.  As Tab.~\ref{table:attr} shows, the improvement comes not only from the addition of auxiliary tasks but also from our approach. This further verifies that our method can be applied to more tasks.

\subsubsection{\textbf{Generalization ability of our method}}
Our method can be generalized to other end-to-end methods. Tab.~\ref{table:gen} shows that our GALW achieves excellent performance across various baselines both on anchor-based method NAE~\cite{NAE} and anchor-free methods AlignPS~\cite{Alignps} and ROI-AlignPS$^\ast$~\cite{Alignps_roi}, which demonstrates the generalization ability of our method.

\setlength{\tabcolsep}{4pt}
\begin{table}[!t]
\begin{center}
\caption{Comparison with state-of-the-art methods on CUHK-SYSU and PRW datasets. Best results are bold in red and the second results are bold in blue.}
\label{table:sota}
\begin{tabular}{c|l|cc|cc}
\hline\noalign{\smallskip}
\multicolumn{2}{c}{\multirow{2}{*}{Methods}}
 & \multicolumn{2}{|c}{CUHK-SYSU} & \multicolumn{2}{|c}{PRW}\\
 \noalign{\smallskip}
\cline{3-6} 
\noalign{\smallskip}
\multicolumn{2}{c|}{} &mAP & Top-1 & mAP & Top-1   \\   
\noalign{\smallskip}
\hline
\noalign{\smallskip}
\multirow{7}{*}{\rotatebox{90}{two-stage}}
&DPM+IDE~\cite{PRW}                & -     & -    & 20.5 & 48.3\\
&CNN+MGTS~\cite{CNN+MGTS}          & 83.3  & 83.9 & 32.8 & 72.1\\
&CNN+CLSA~\cite{CNN+CLSA}          & 87.2  & 88.5 & 38.7 & 65.0\\
&FPN+RDLR~\cite{FPN+RDLR}          & \textbf{\textcolor{blue}{93.0}}  & \textbf{\textcolor{blue}{94.2}} & 42.9 & 70.2\\
&IGPN~\cite{IGPN}                  & 90.3  & 91.4 & \textbf{\textcolor{blue}{47.2}} & \textbf{\textcolor{blue}{87.0}}\\
&OR~\cite{OR}                      & 92.3  & 93.8 &\textbf{\textcolor{red}{52.3}} & 71.5\\
&TCTS~\cite{TCTS}                  & \textbf{\textcolor{red}{93.9}}  & \textbf{\textcolor{red}{95.1}} & 46.8 & \textbf{\textcolor{red}{87.5}}\\
\hline 
\hline 
\multirow{20}{*}{\rotatebox{90}{one-stage}}
&OIM~\cite{CUHK}                   & 75.5  & 78.7 & 21.3 & 49.4\\
&NPSM~\cite{NPSM}                  & 77.9  & 81.2 & 24.2 & 53.1\\
&RCAA~\cite{RCAA}                  & 79.3  & 81.3 & -    & -   \\
&CTXG~\cite{CTXG}                  & 84.1  & 86.5 & 33.4 & 73.6\\
&QEEPS~\cite{QEEPS}                & 88.9  & 89.1 & 37.1 & 76.7\\
&HOIM~\cite{HOIM}                  & 89.7  & 90.8 & 39.8 & 80.4\\
&BINet~\cite{BINet}                & 90.0  & 90.7 & 45.3 & 81.7\\
&NAE~\cite{NAE}                    & 91.5  & 92.4 & 43.3 & 80.9\\
&PGA~\cite{PGA}                    & 92.3  & 94.7 & 44.2 & 85.2\\
&SeqNet~\cite{SeqNet}              & 93.8  & 94.6 & 46.7 & 83.4\\
&AGWF~\cite{TriNet}              & 93.3  & 94.2 & \textbf{\textcolor{red}{53.3}} &\textbf{\textcolor{red}{87.7}} \\
&AlignPS~\cite{Alignps}            & 93.1  & 93.4 & 45.9 & 81.9\\
&ROI-AlignPS~\cite{Alignps_roi}    & \textbf{\textcolor{blue}{95.4}}  & \textbf{\textcolor{blue}{96.0}} & 51.6 & 84.4\\
&ROI-AlignPS$^\ast$~\cite{Alignps_roi}    & 95.0  & 95.3 & 50.3 & 84.3\\

\cline{2-6} 
\noalign{\smallskip}
&ROI-AlignPS$^\ast$ w/ GALW    & \textbf{\textcolor{red}{95.6}} & \textbf{\textcolor{red}{96.3}} & \textbf{\textcolor{blue}{52.9}} & \textbf{\textcolor{blue}{86.1}}\\   
\noalign{\smallskip}
\hline
\end{tabular}
\end{center}
\end{table}
\setlength{\tabcolsep}{1.4pt}

\begin{figure}[!t]
\setlength{\abovecaptionskip}{1mm} 
  \centering
  \includegraphics[width=0.9\linewidth]{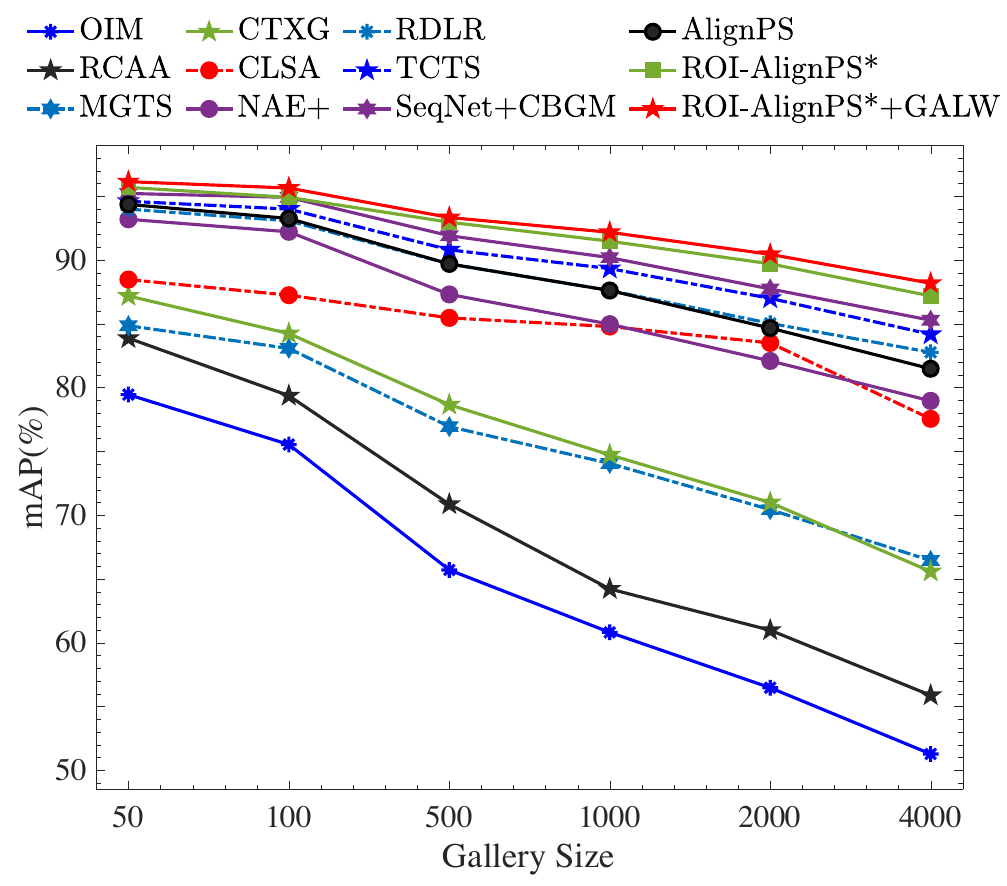}
  \caption{The mAP under different gallery sizes on CUHK-SYSU dataset. The dashed lines represent two-stage methods and the solid lines represent one-stage ones.}
  \Description{...}
  \label{fig:gallery}
\end{figure}

\setlength{\tabcolsep}{4pt}
\begin{table}[!t]
\begin{center}
\caption{Speed comparison on different GPUs. Runtimes are measured in milliseconds.}
\label{table:runtime}
\begin{tabular}{l|cc}
\hline\noalign{\smallskip}
\multirow{2}{*}{Methods}& \multicolumn{2}{c}{GPU (TFLOPs)} \\
\noalign{\smallskip}
\cline{2-3}  \noalign{\smallskip}
& P40(11.8) & RTX(16.3)\\
\noalign{\smallskip}
\hline
\noalign{\smallskip}
NAE~\cite{NAE}                    & 158  & 85 \\
AlignPS~\cite{Alignps}            & 122  & 65\\
SeqNet~\cite{SeqNet}              & 178  & 97\\ 
AGWF~\cite{TriNet}              & 145  & 80\\ 
ROI-AlignPS$^\ast$~\cite{Alignps_roi}        & \textbf{127}  & \textbf{69}\\
ROI-AlignPS$^\ast$ w/ GALW        & \textbf{127}  & \textbf{69}\\

\noalign{\smallskip}
\hline
\end{tabular}
\end{center}
\end{table}
\setlength{\tabcolsep}{1.4pt}

\begin{figure*}[ht]
\setlength{\abovecaptionskip}{0.88mm} 
  \centering
  \includegraphics[width=0.9\linewidth]{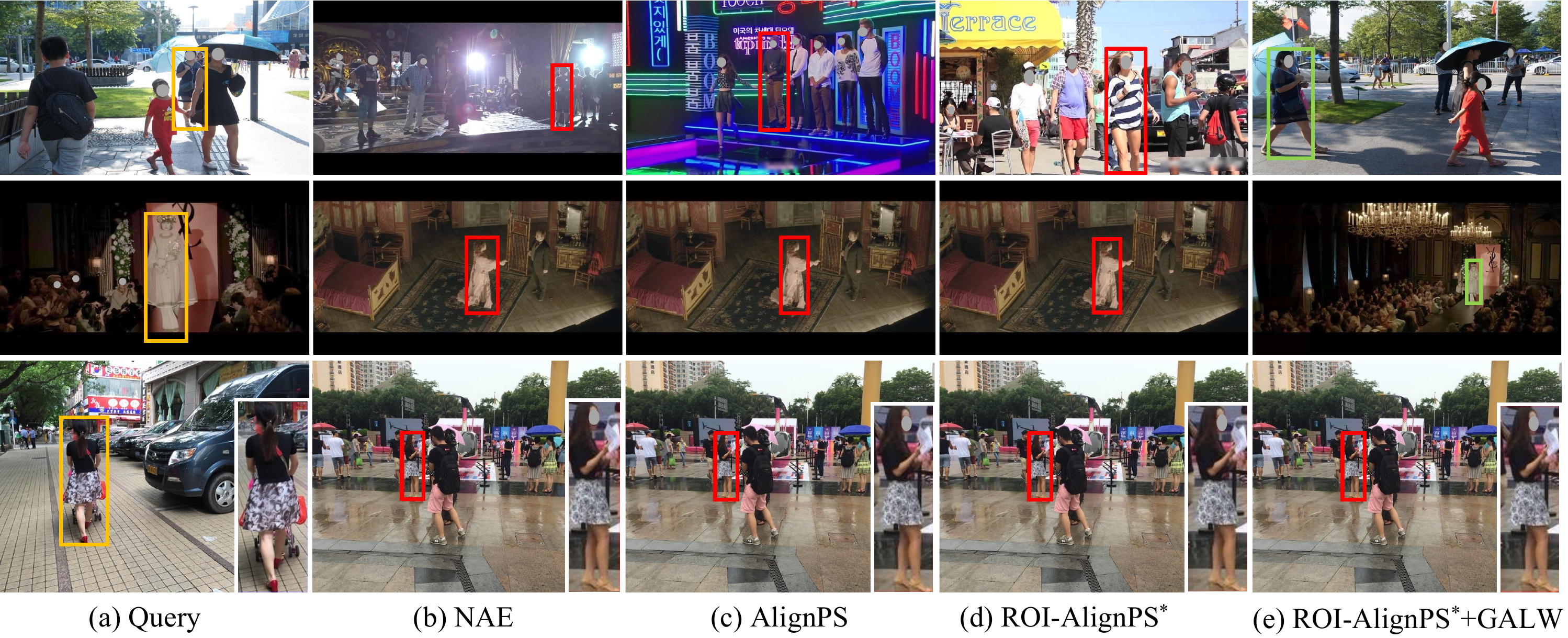}
  \caption{Top-1 search results for several samples. The orange, green and red bounding boxes denote the queries, correct and incorrect matches, respectively. A failure case is in the last row and we zoom the detected object for better view.}
  \Description{...}
  \label{fig:qualitative}
\end{figure*}

\subsection{Comparison to the State-of-the-Art Methods}
In Tab.~\ref{table:sota}, we compare our method with the state-of-the-art methods, including two-stage methods~\cite{PRW,CNN+MGTS,CNN+CLSA,FPN+RDLR,IGPN,OR,TCTS} and one-stage methods~\cite{CUHK,NPSM,RCAA,CTXG,QEEPS,HOIM,BINet,NAE,PGA,SeqNet,TriNet,Alignps,Alignps_roi}.

\subsubsection{Results on CUHK-SYSU dataset.}
The performance with our proposed method on CUHK-SYSU dataset~\cite{CUHK} is 95.6\% and 96.3\% in terms of mAP and Top-1 scores, respectively. Notably, whether compared with the one-stage methods or the two-stage methods, our method achieves the best performance, which is 0.6 and 1.0 pp. higher w.r.t. mAP and Top-1 than the baseline ROI-AlignPS$^\ast$.

In addition, we further evaluate the performance under larger search scopes. {\it i.e.} each query person is matched in galleries of different sizes. From Fig.~\ref{fig:gallery}, we can see that the mAPs for all methods decrease monotonically with increasing gallery sizes, which means it is difficult to match a person in larger scopes. We can observe that the framework with our method GALW outperforms both one-step methods and two-step methods at all gallery sizes.

\subsubsection{Results on PRW dataset.}
On PRW dataset~\cite{PRW}, all methods suffer from degraded performance due to the characteristic of fewer training images and a larger gallery. However, the proposed method outperforms AGWF~\cite{TriNet}, which is the current state-of-the-art one-stage method by 53.3\% in mAP. Our method improves over baseline by 2.6 and 1.8 pp. w.r.t. mAP and top-1 scores. This margin means our method is also robust on small datasets.

\subsubsection{Runtime Comparison.}
We compare the speed of different models in a P40 and RTX GPU respectively. All methods are implemented in PyTorch~\cite{pytorch} without bells and whistles. We test inference time with input images in size 1500$\times$900. As shown in Tab.~\ref{table:runtime}, ROI-AlignPS$^\ast$ with our method cost 127 and 69 milliseconds on a P40 and a RTX GPU respectively. Our method is faster than AGWF~\cite{TriNet}, which is the current state-of-the-art one-step method, while achieving competitive performance on PRW dataset.

\subsubsection{Qualitative Results.}
We show some qualitative search results of NAE~\cite{NAE}, AlignPS~\cite{Alignps}, ROI-Alignps$^\ast$+GALW and ROI-AlignPS$^\ast$~\cite{Alignps_roi} in Fig.~\ref{fig:qualitative}. We can see that when we apply our proposed GALW to ROI-AlignPS$^\ast$, it can further successfully handle cases including occlusions (row 1), viewpoint variation (row 1, 2) and scale variation (row 2). A failure case is illustrated in the last row, where it still fails in distinguishing objects that share similar appearances.

\section{Conclusions}
In this paper, we propose a GALW for person search. It is a great challenge how to weight the tasks automatically and dynamically, especially with a large number of different tasks in one-stage methods of person search. Since person search is a typical multi-task problem and loss weighting is a straightforward way to resolve it, we try to make an analysis on different baselines with an existing loss weighing method. From the analysis, we find it under-performs with a large number of tasks and the issue of inconsistent convergence rates gets severe over increasing tasks. Motivated by these findings, we present a GALW method, grouping tasks into a group according to convergence rate andassigning each task group with a learnable loss weight, which makes training of person search more effective. In addition, in order to further verify the generalization ability and extendability of GALW with more tasks, we apply GALW on different baselines and introduce an attribute recognition to our baseline network in Fig.~\ref{fig:network} as an auxiliary task respectively. Experimental results demonstrate that our method can weight tasks more effectively and it is still valid for more tasks.

\begin{acks}
This work was supported in part by the National Natural Science Foundation of China (Grant No. 62172225) and the Fundamental Research Funds for the Central Universities (No. 30920032201).
\end{acks}

\bibliographystyle{ieee_fullname}
\bibliography{sample-base}

\appendix









\end{document}